\documentclass[lettersize,journal]{IEEEtran}
\usepackage{amsmath,amsfonts}
\usepackage{algorithmic}
\usepackage{algorithm}
\usepackage{array}
\usepackage[caption=false,font=normalsize,labelfont=sf,textfont=sf]{subfig}
\usepackage{textcomp}
\usepackage{stfloats}
\usepackage{url}
\usepackage{verbatim}
\usepackage{graphicx}
\usepackage{cite}
\usepackage{booktabs}
\usepackage{tabularx}
\usepackage{caption}
\usepackage{soul}
\usepackage{pifont}
\usepackage{multirow}
\usepackage{threeparttable}
\usepackage{makecell}
\usepackage{graphicx}
\usepackage{siunitx} 
\newcommand{\cmark}{\ding{51}}

\begin{document}
\raggedbottom                                                                                                                                                                                                                                                                                              
\title{Physfusion: A Transformer-based Dual-Stream Radar and Vision Fusion Framework for Open Water Surface Object Detection}

\author{Yuting Wan, Liguo Sun, Jiuwu Hao, Zao Zhang, Pin LV~\IEEEmembership{Member,~IEEE,}
\thanks{Yuting Wan and Jiuwu Hao are with the School of Artificial Intelligence, University of Chinese Academy of Sciences, Beijing, China.}%
\thanks{Liguo Sun, Zao Zhang and Pin Lv are with the Institute of Automation, Chinese Academy of Sciences, Beijing, China.}%

}

\markboth{Journal of \LaTeX\ Class Files,~Vol.~14, No.~8, August~2021}%
{Shell \MakeLowercase{\textit{et al.}}: A Sample Article Using IEEEtran.cls for IEEE Journals}


\maketitle

\begin{abstract}
Detecting water-surface targets for Unmanned Surface Vehicles (USVs) is challenging due to non-stationary wave clutter, specular reflections, and frequent long-range observations where small or low-profile objects exhibit weak appearance cues. Although 4D millimeter-wave radar can complement cameras under degraded illumination, maritime radar point clouds are typically sparse and intermittent, and reflectivity-related attributes often present heavy-tailed variations under scattering and multipath, making it difficult for conventional fusion designs to exploit radar cues effectively and maintain stable predictions across frames.

We propose PhysFusion, a physics-informed radar--image detection framework tailored for water-surface perception. The framework integrates: (1) a Physics-Informed Radar Encoder (PIR Encoder) composed of an RCS Mapper and a Quality Gate, which transforms per-point radar attributes into a compact scattering prior and predicts point-wise reliability to support robust feature learning under clutter; (2) a Radar-guided Interactive Fusion Module (RIFM) that performs query-level radar--image fusion between semantically enriched radar features and multi-scale visual features, with the radar branch modeled by a dual-stream backbone including a point-based local stream and a transformer-based global stream using Scattering-Aware Self-Attention (SASA); and (3) a Temporal Query Aggregation module (TQA) to aggregate frame-wise fused queries over a short temporal window and obtain temporally consistent representations for the detection head. Experiments on WaterScenes and FLOW demonstrate that PhysFusion achieves 59.7\% mAP$_{50:95}$ and 90.3\% mAP$_{50}$ on WaterScenes (with $T{=}5$ radar history) using 5.6M parameters and 12.5G FLOPs, and reaches 94.8\% mAP$_{50}$ and 46.2\% mAP$_{50:95}$ on FLOW under the radar+camera setting. Ablation studies on WaterScenes further quantify the contributions of the PIR Encoder, SASA-based global radar reasoning, and RIFM under a consistent evaluation protocol.
\end{abstract}


\section{Introduction}
\IEEEPARstart{P}{erception} for Unmanned Surface Vehicles (USVs) presents challenges distinct from terrestrial autonomous driving. In addition to dynamic water-surface reflections and wave-induced non-stationary backgrounds, USV environments often contain small, low-profile objects (e.g., buoys, kayaks, floating debris) and involve long-range observations where visual cues can be weak or ambiguous. These conditions can reduce the reliability of purely vision-based detection, particularly under adverse weather or lighting such as fog, rain, glare, or low-light situations~\cite{10965796, yao2024waterscenes, wang2024mdd, qu2024double}.

4D millimeter-wave radar provides complementary sensing information, offering range, velocity, and reflectivity-based intensity measurements that are less sensitive to illumination and visibility changes~\cite{bilik2022comparative}. However, effectively fusing radar with camera data in aquatic environments remains a challenge. Existing radar-camera fusion frameworks, primarily developed for road scenes, often treat radar as a secondary geometric input and employ fixed fusion strategies (e.g., early concatenation or late fusion). These approaches may not adequately address the inherent heterogeneities between sparse, physics-rich radar point clouds and dense, appearance-based image features~\cite{guan2023achelous, zhu2023millimeter}. Specific limitations include: (i) underutilization of the semantic information embedded in radar intensity for material discrimination, (ii) lack of explicit, learnable mechanisms to align the two modalities and resolve spatial mismatches, and (iii) sensitivity to intermittent radar returns caused by wave occlusion and platform motion.

To address these limitations, we propose PhysFusion, a radar-image fusion framework designed for water-surface object detection. The framework is built around a physics-informed radar encoder and a temporal query aggregation mechanism, aiming to achieve semantically guided and spatiotemporally aligned fusion. Our main contributions are threefold:

\begin{itemize}
\item PIR Encoder (Physics-Informed Radar Encoder). We introduce a physics-informed radar encoder that explicitly leverages reflectivity-related attributes together with geometry. The PIR Encoder consists of an \emph{RCS Mapper} that transforms per-point radar attributes into a compact scattering prior and a \emph{Quality Gate} that predicts point-wise reliability, improving robustness to heavy-tailed intensity variations and intermittent returns in maritime radar.

\item RIFM (Radar-guided Interactive Fusion Module). We propose a radar-guided interactive fusion module that performs query-level radar--image fusion. Semantically enriched radar features are used to attend to and aggregate relevant multi-scale visual features, enabling adaptive alignment between sparse radar evidence and dense image regions. In addition, the radar branch adopts a dual-stream backbone with a point-based local stream and a transformer-based global stream equipped with Scattering-Aware Self-Attention (SASA) for global reasoning over sparse points.

\item TQA-GRU (Temporal Query Aggregation). We introduce temporal query aggregation to improve frame-to-frame continuity under platform motion and intermittent observations. TQA-GRU aggregates fused cross-modal queries over a short temporal window, implemented with a lightweight GRU, to obtain temporally consistent representations for set-prediction detection.
\end{itemize}

Experiments on the WaterScenes~\cite{yao2024waterscenes} and FLOW~\cite{Cheng_2021_ICCV} datasets indicate that PhysFusion can achieve competitive 2D detection accuracy while maintaining inference efficiency suitable for edge deployment. Ablation studies examine the contributions of the PIR Encoder, the RIFM-based fusion design, and TQA, as well as the role of SASA for global radar modeling under water-surface sensing conditions.
\section{Related Work}
\label{sec:related_work}

\subsection{Multi-Modal Fusion for Object Detection}
Multi-modal fusion is widely used in autonomous perception to combine complementary cues from RGB cameras and active sensors such as LiDAR and millimeter-wave radar. Existing fusion pipelines are commonly categorized into early (sensor-level), middle (feature-level), and late (decision-level) fusion strategies~\cite{Vora2019PointPaintingSF, Bai2022TransFusionRL, liu2023bevfusion}. 
Early fusion typically projects measurements into a shared representation (e.g., image plane or BEV) and then applies a unified backbone, while late fusion aggregates modality-specific predictions. Middle fusion has attracted increasing attention because it can preserve modality-specific inductive biases while enabling cross-modal interaction~\cite{cai2023bevfusion4d,  chang2025recurrentbev, liu2023bevfusion}, and attention/Transformer-based designs are frequently adopted to model long-range dependencies and soft correspondence across modalities~\cite{Bai2022TransFusionRL, liu2023bevfusion, Liang2022BEVFusionAS}. 

Most of these methods are developed for structured road scenes where sensor returns and background statistics are comparatively stable. In contrast, water-surface environments introduce non-stationary clutter (e.g., waves and reflections), long-range observation regimes with weak appearance cues, and modality-specific degradations, which can complicate cross-modal alignment and feature aggregation.

\subsection{Radar-Based Perception and Radar--Camera Fusion}
Compared with LiDAR, millimeter-wave radar provides range and Doppler velocity and is less affected by illumination changes, making it a complementary modality under low visibility~\cite{ouaknine2021carrada, bijelic2020seeing, bilik2022comparative}. 
However, radar point clouds are typically sparse, and their reflectivity-related attributes (e.g., RCS or intensity proxies) can vary significantly due to scattering, multipath, and viewpoint effects. This often limits the effectiveness of directly treating reflectivity as a raw ``intensity channel'' and may introduce sensitivity to clutter and heavy-tailed distributions in practice~\cite{cheng2021flow}. 

Radar--camera fusion has therefore been explored to combine radar's geometric/kinematic cues with image semantics. Representative approaches differ by how they align heterogeneous measurements, e.g., projecting radar points to the image plane, performing feature concatenation, or applying attention to learn soft correspondences~\cite{chadwick2019distant, nabati2021centerfusion, kim2023craft, zhang2023cmx}. 
While these designs are effective under certain settings, their performance can be influenced by the stability of radar returns and the quality of cross-modal alignment, particularly when radar observations are sparse or contain substantial clutter.

\subsection{Physics-Informed Modeling for Radar Features}
A growing line of research incorporates domain knowledge or physically motivated priors into perception models to improve feature learning under sensor noise and environment-induced artifacts. This trend is particularly relevant to radar-based perception, where measurements encode not only geometry but also motion and scattering characteristics, such as range, radial velocity, and reflectivity proxies (e.g., RCS)~\cite{bilik2022comparative}. In contrast to LiDAR, radar observations are often sparse, spatially non-uniform, and subject to multipath and specular scattering, which can induce heavy-tailed statistics and frame-to-frame intermittency in reflectivity-related channels~\cite{ouaknine2021carrada}. These properties make it suboptimal to treat reflectivity as a raw ``intensity'' feature without calibration or uncertainty awareness, especially in cluttered maritime scenes.

A common strategy is therefore to encode radar attributes with representations that are more stable and semantically meaningful than raw channels. In road-scene radar--camera fusion, several works exploit Doppler and reflectivity cues to enhance detection and cross-modal association~\cite{nabati2021centerfusion,kim2023craft,lin2024rcbevdet}. Related efforts also emphasize learned attribute mapping, normalization, or feature modulation to reduce sensitivity to sensor-dependent scaling and heavy-tailed distributions~\cite{10225711,11127186}. In parallel, reliability-aware modeling has been explored to mitigate the influence of spurious returns: a point-wise confidence (or uncertainty) estimate can gate feature aggregation so that transient clutter points contribute less to neighborhood pooling or attention computation. Such designs are aligned with physics-informed encoders that introduce learned priors or gating mechanisms to suppress clutter/multipath effects while preserving informative scattering cues for targets~\cite{lin2024rcbevdet, 10225711,11127186}. In the context of water-surface perception, these ideas motivate mapping raw radar attributes to compact scattering descriptors and explicitly modeling point-level reliability before downstream local/global aggregation.

\subsection{Transformer Attention for Sparse Point Sets}
Transformers have been widely adopted for point-based representations due to their capacity for long-range dependency modeling and flexible set processing~\cite{Bai2022TransFusionRL}. In 3D perception, Transformer variants have demonstrated strong capability in capturing contextual relations among points, voxels, or regions~\cite{zhao2021pointtransformer,mao2021votr}. However, directly applying standard global self-attention to sparse point sets can be problematic. When observations are sparse and noisy, attention weights driven primarily by feature similarity may overemphasize spatially distant but feature-similar points, leading to topology-inconsistent aggregation and susceptibility to clutter-induced correlations.
\begin{figure*}[ht]

        \includegraphics[width=\linewidth]{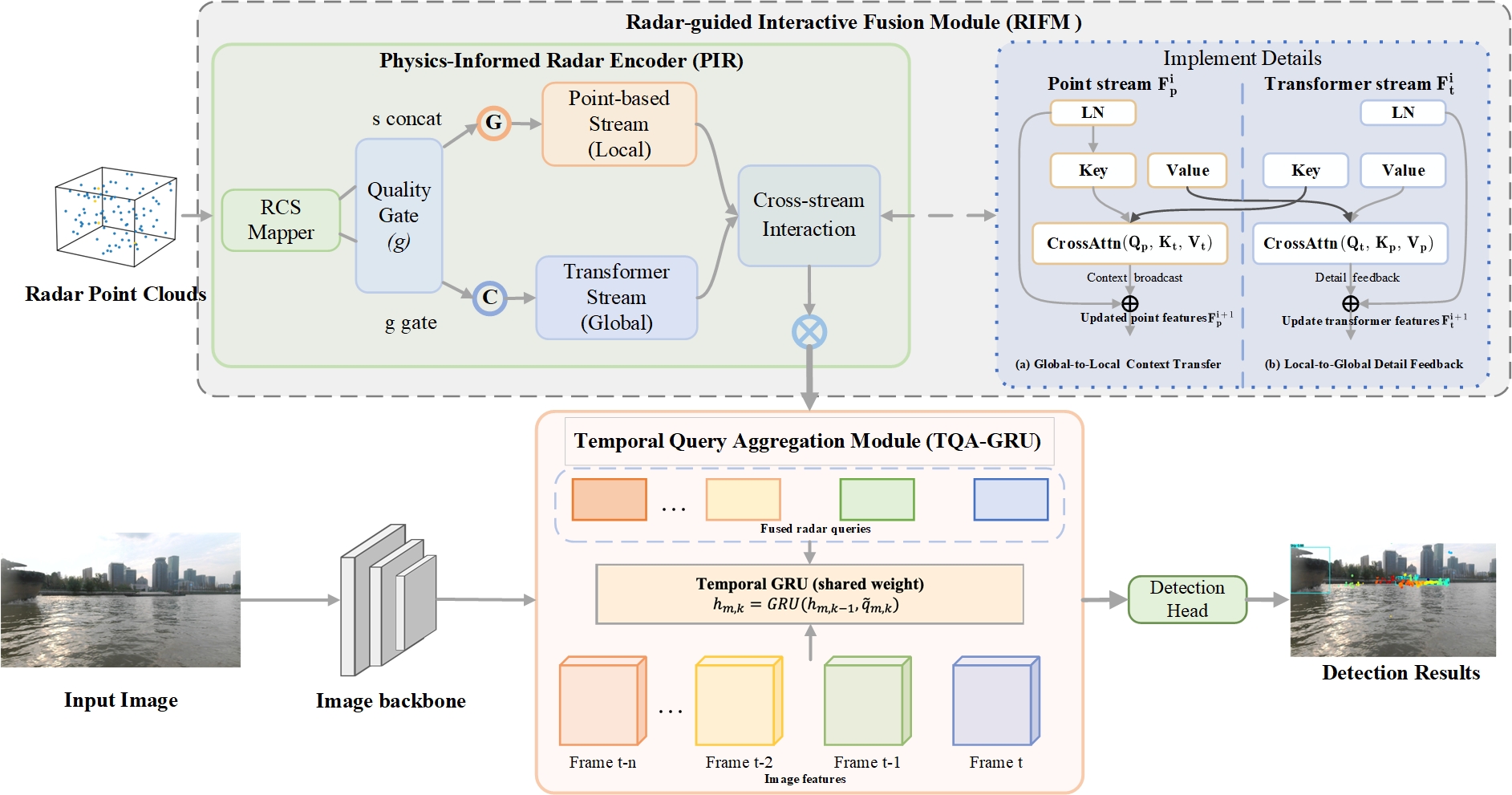}
\caption{Overview of PhysFusion. The framework consists of three contribution-aligned modules: (i) PIR Encoder, where an RCS Mapper produces a compact scattering prior $s$ and a Quality Gate predicts a point-wise reliability score $g$ to modulate radar features under heavy-tailed reflectivity and intermittent returns; (ii) RIFM, which adopts a dual-stream radar backbone with a point-based local stream and a Transformer-based global stream equipped with SASA (Scattering-Aware Self-Attention), together with cross-stream interaction to exchange local details and global context, and performs query-level radar--image fusion; and (iii) TQA-GRU, which aggregates per-frame fused cross-modal queries over a temporal window using shared weights to obtain temporally consistent query representations. The aggregated queries are finally fed into a detection head to output 2D water-surface object detections.}
    \label{fig:main}
\end{figure*}
To alleviate these issues, many 3D designs incorporate inductive biases that encourage locality, such as restricted attention windows, sparse attention patterns, or distance-aware weighting, thereby promoting physically plausible interactions~\cite{zhao2021pointtransformer,mao2021votr}. A representative mechanism is distance-decayed attention, which modifies attention logits by penalizing pairs with large spatial separation. This formulation reduces the chance of forming spurious long-range associations while retaining meaningful interactions among nearby points. Related ideas also appear in deformable/sparse attention paradigms that reduce computation and concentrate attention on informative neighborhoods or sampled locations~\cite{Zhao_2024_CVPR,zhu2021deformable}. For radar point sets, these considerations are especially important because sparsity and clutter similarity can otherwise dominate global reasoning. Consequently, introducing scattering-/distance-aware priors into attention is a principled way to stabilize Transformer-based global modeling under sparse radar returns.

\subsection{Temporal Modeling for Intermittent Observations}
Temporal aggregation is a common strategy to improve prediction stability when single-frame observations are incomplete or noisy. Prior work leverages recurrent models (e.g., GRU/LSTM), temporal attention, or multi-frame feature fusion to accumulate evidence across time for detection and tracking~\cite{cai2023bevfusion4d,   chang2025recurrentbev, WAN2026131914}. For transformer-based detection and tracking, query-centric temporal modeling has become an effective paradigm: instead of densely accumulating features, the model maintains a fixed set of object queries (or track queries) and updates them across frames, which provides a structured way to preserve identity and temporal continuity~\cite{Li2022BEVFormerLB, 10965796}.

This perspective is particularly suitable for radar and water-surface scenarios where the number of returns and their spatial distributions can vary substantially across frames due to wave motion, multipath, and intermittent target visibility. Query-based temporal modeling converts frame-wise features into a compact, fixed-size representation and then aggregates query states over time, making it less sensitive to variable point counts than direct feature stacking. Moreover, temporal modeling can be performed at different stages: (i) early fusion across raw measurements, (ii) feature-level temporal fusion, or (iii) query-level temporal aggregation for set prediction. For USV perception, the latter is attractive because it decouples temporal consistency from per-frame sparsity and enables temporally consistent predictions even when radar evidence is missing in some frames. These observations motivate temporal query aggregation designs (e.g., GRU-based updates with shared weights across queries) that maintain consistent query identities while improving frame-to-frame continuity of detection outputs.

\section{Methodology}
\label{sec:method}

Perception in water-surface environments is substantially more challenging than land-based autonomous driving due to dynamic clutter, adverse illumination, and frequent sensing degradation. In particular, 4D millimeter-wave radar returns on water are often sparse and intermittent under wave motion, water vapor, specular reflections, and long-range scattering, producing noisy point clouds with unstable reflectivity patterns (RCS) and missing returns. These characteristics make robust detection difficult if the model relies on purely data-driven feature similarity or single-frame cues.

To address these issues, we propose PhysFusion, As illustrated in Fig.~\ref{fig:main}, a physics-informed radar--image fusion framework designed for water-surface object detection. PhysFusion is organized around three components aligned with our contributions: (i) a Physics-Informed Radar Encoder (PIR Encoder) that stabilizes radar representations under clutter and heavy-tailed RCS, (ii) a Radar-guided Interactive Fusion Module (RIFM) that performs query-level radar--image fusion with radar-guided cross-modal interaction, and (iii) Temporal Query Aggregation (TQA) that aggregates fused cross-modal queries over a short temporal window to improve frame-to-frame continuity. Within the radar branch, we adopt a \emph{dual-stream backbone} comprising a point-based \emph{local stream} and a Transformer-based \emph{global stream}. The global stream uses SASA (Scattering-Aware Self-Attention) to bias global reasoning toward physically plausible interactions over sparse radar points.
\begin{figure*}[ht]

        \includegraphics[scale=0.7]{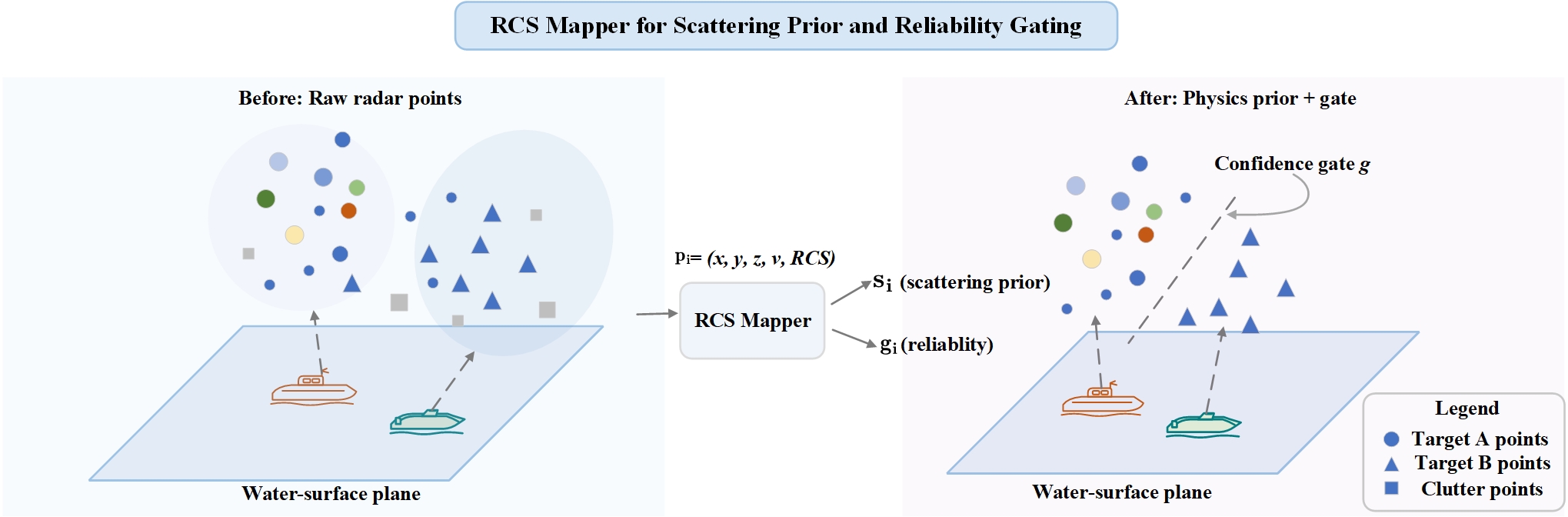}
\caption{Illustration of the PIR Encoder with the RCS Mapper and Quality Gate. Left: raw radar returns on the water-surface plane include target points and clutter/outliers. Middle: each point $p_i=(x_i,y_i,z_i,v_i,\mathrm{RCS}_i)$ is mapped to a compact scattering prior $s_i$ and a point-wise reliability score $g_i$. Right: the scattering prior augments radar features, while the confidence gate down-weights unreliable returns (e.g., heavy-tailed or intermittent scattering), yielding a reliability-aware radar representation for subsequent encoding and fusion.}

    \label{fig:RCS-mapper}
\end{figure*}
\subsection{PIR Encoder: Physics-Informed Radar Encoding}
Water-surface 4D radar returns are often sparse and intermittent, and their reflectivity attributes (e.g., RCS) can exhibit heavy-tailed fluctuations due to scattering and clutter. As shown in Fig.~\ref{fig:RCS-mapper}, PIR Encoder learns a compact scattering prior with an RCS Mapper and predicts a point-wise reliability score with a Quality Gate to down-weight unreliable returns before feature aggregation.

\label{subsec:pir}
\paragraph{Radar point representation.}
Given the sparsity and intermittency of water-surface radar returns, we represent each frame as an unordered set of radar points and encode each point with both geometry and reflectivity-related attributes. Specifically, the input radar point cloud is
\begin{equation}
P = \{ p_i \}_{i=1}^N, \quad p_i = (x_i, y_i, z_i, v_i, \mathrm{RCS}_i),
\end{equation}
where $(x_i,y_i,z_i)$ denotes the 3D coordinates, $v_i$ is the Doppler radial velocity, and $\mathrm{RCS}_i$ is the radar cross section. For convenience, we denote $\mathbf{x}_i=(x_i,y_i,z_i)$. 
Before feature learning, we apply a lightweight normalization to reduce scale mismatch across attributes and to stabilize optimization. We then construct a compact attribute vector
\paragraph{RCS Mapper: compact scattering prior.}
Directly using raw $\mathrm{RCS}_i$ can be sensitive to heavy-tailed variations induced by scattering and multipath. 
We introduce an \emph{RCS Mapper} as illustrated in Fig.~\ref{fig:RCS-mapper} that transforms per-point radar attributes into a compact \emph{scattering prior}:

\begin{equation}
s_i=\mathcal{M}(a_i;\Theta)=\mathrm{MLP}_{\mathrm{map}}(a_i;\Theta)\in\mathbb{R}^{C_s}.
\end{equation}
Compared with feeding $\mathrm{RCS}_i$ directly, $s_i$ serves as a learnable bounded descriptor that is less sensitive to heavy-tailed reflectivity fluctuations while still retaining reflectivity-related cues.

\paragraph{Quality Gate: point-wise reliability.}
In parallel, a \emph{Quality Gate} predicts point-wise reliability $g_i\in(0,1)$:
\begin{equation}
g_i = \sigma\!\left(\mathrm{MLP}_{\mathrm{gate}}([a_i;\,s_i])\right),
\end{equation}
where $\sigma(\cdot)$ is the sigmoid function. The gate modulates the point embedding so that unreliable/intermittent returns contribute less:
\begin{equation}
f_i^{(0)}=\mathrm{MLP}_{\mathrm{in}}([a_i;\,s_i]),\qquad
\tilde{f}_i^{(0)} = g_i\cdot f_i^{(0)}.
\end{equation}
Together, the RCS Mapper and Quality Gate constitute the PIR Encoder, providing a physics-informed prior and reliability-aware filtering for maritime radar point clouds.
\begin{figure}[!t]
    \centering
    \includegraphics[width=\linewidth]{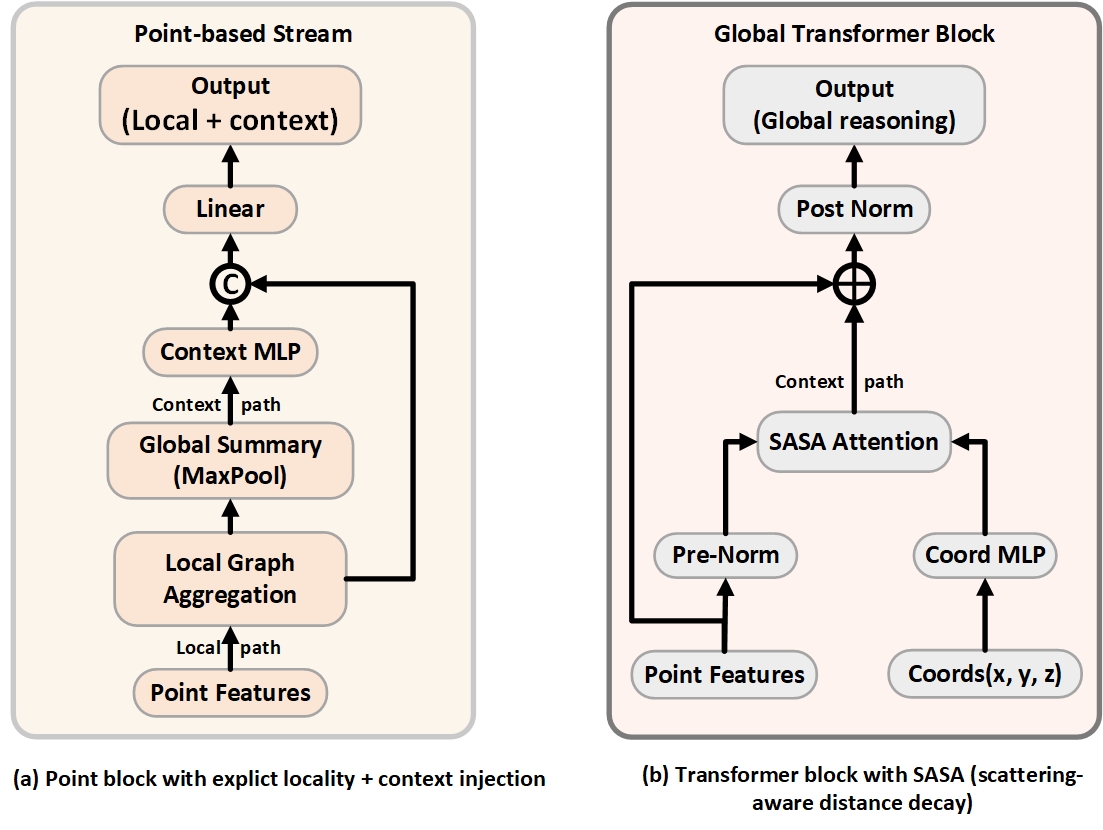}
\caption{Dual-stream radar backbone for radar point encoding. (a) Point-based local stream performs local graph aggregation and injects global context via a pooled summary and a lightweight context MLP, producing locality-preserving features with contextual cues. (b) Global Transformer block for radar modeling, where SASA incorporates a distance-decay prior from point coordinates to suppress spurious long-range interactions and improve global reasoning over sparse returns.}

    \label{fig:dual-stream}
\end{figure}
\subsection{RIFM: Radar-guided Interactive Fusion Module}
\label{subsec:rifm}
Water-surface radar point clouds are sparse and irregular, with weak target clusters mixed with clutter. Local-only aggregation may miss broader context, while global-only attention can introduce spurious long-range associations. To balance both, RIFM adopts a dual-stream radar backbone that combines local neighborhood modeling with global reasoning to produce semantically enriched radar tokens for query-level radar--image fusion, as shown in Fig.~\ref{fig:dual-stream}.

\paragraph{Dual-stream radar backbone overview.}
Given PIR-modulated features $\{\tilde{f}_i^{(0)}\}$, we extract radar representations with a dual-stream backbone, as shown in Fig.~\ref{fig:dual-stream}:
(i) a point-based \emph{local stream} emphasizing neighborhood structure, and
(ii) a Transformer-based \emph{global stream} for long-range reasoning over sparse points.
Let the resulting local/global features be $\{f^{\mathrm{loc}}_i\}$ and $\{f^{\mathrm{glo}}_i\}$, which are fused into radar tokens
\begin{equation}
f_i^{r}=\mathrm{Proj}\big([f_i^{\mathrm{loc}};\,f_i^{\mathrm{glo}}]\big),\qquad
F^{r}=\{f_i^{r}\}_{i=1}^{N}.
\end{equation}

\paragraph{Point-based local stream (physics-aware kNN + dynamic graph aggregation).}
To emphasize locality, for each point $p_i$ we construct a neighborhood $\mathcal{N}(i)$ using kNN under a physics-aware metric:
\begin{equation}
d_{ij}^2 = \| \mathbf{x}_i - \mathbf{x}_j \|_2^2 + \lambda_v (v_i - v_j)^2 + \lambda_r (\mathrm{RCS}_i - \mathrm{RCS}_j)^2,
\end{equation}
where $\lambda_v,\lambda_r$ control the contributions of Doppler and RCS similarity.
We aggregate local edge features following dynamic graph convolution~\cite{wang2019dgcnn}:
\begin{equation}
\resizebox{\linewidth}{!}{$
e_{ij}^{(s)}=\phi^{(s)}\!\left(\Big[\tilde{f}_i^{(s)},\, \tilde{f}_j^{(s)},\, (\mathbf{x}_j-\mathbf{x}_i),\, (v_j-v_i),\, (\mathrm{RCS}_j-\mathrm{RCS}_i)\Big]\right),\quad j\in\mathcal{N}(i).
$}
\end{equation}

\begin{equation}
\tilde{f}_i^{(s+1)}=\max_{j\in\mathcal{N}(i)}\left(e_{ij}^{(s)}\right),
\end{equation}
where $\phi^{(s)}(\cdot)$ is a shared MLP and $\tilde{f}_i^{(s)}$ denotes the point feature at layer $s$ (with PIR gating applied before and/or within local aggregation).
This local inductive bias helps preserve weak target clusters while suppressing isolated clutter outliers.

\paragraph{Transformer-based global stream with SASA.}
The global stream models broader dependencies among scattered points.
Applying standard self-attention to sparse radar points can introduce topology-inconsistent associations due to the lack of an explicit physical bias.
We adopt SASA by adding a soft distance-decay prior to attention logits:
\begin{equation}
\mathrm{SASA}(Q,K,V)=\mathrm{Softmax}\!\left(\frac{QK^{\top}}{\sqrt{d_k}}-\beta D^2\right)V,
\end{equation}
where $Q,K,V$ are the query/key/value projections of radar tokens, $d_k$ is the key dimension, $D$ is the Euclidean distance matrix between points, and $\beta$ is a learnable decay parameter.
SASA encourages attention to concentrate on physically plausible interactions under sparse and noisy returns.

\paragraph{Query-level radar--image fusion.}
Let the image backbone output multi-scale features $\{F^{img}_\ell\}_{\ell=1}^{L}$, and define a flattened visual token set
\begin{equation}
F^{img} = \mathrm{Flatten}\big(\{F^{img}_\ell\}_{\ell=1}^{L}\big).
\end{equation}
RIFM performs fusion at the \emph{query level} using $M$ learnable object queries $\{q_m\}_{m=1}^{M}$ (set-prediction style).
For each query, we extract modality-specific query features via cross-attention:
\begin{equation}
\begin{aligned}
q^{r}_{m} &= \mathrm{CrossAttn}\!\big(\mathrm{LN}(q_m),\ \mathrm{LN}(F^{r})\big),\\
q^{img}_{m} &= \mathrm{CrossAttn}\!\big(\mathrm{LN}(q_m),\ \mathrm{LN}(F^{img})\big).
\end{aligned}
\end{equation}

and fuse them with an MLP:
\begin{equation}
\tilde{q}_{m}=\mathrm{MLP}_{\mathrm{fuse}}\big([q^{r}_{m};\,q^{img}_{m}]\big).
\end{equation}

Intuitively, semantically enriched radar representations (stabilized by PIR and modeled by the dual-stream backbone) guide the aggregation of relevant multi-scale visual evidence, enabling adaptive alignment between sparse radar cues and dense image regions.
\begin{figure*}[ht]

        \includegraphics[width=\linewidth]{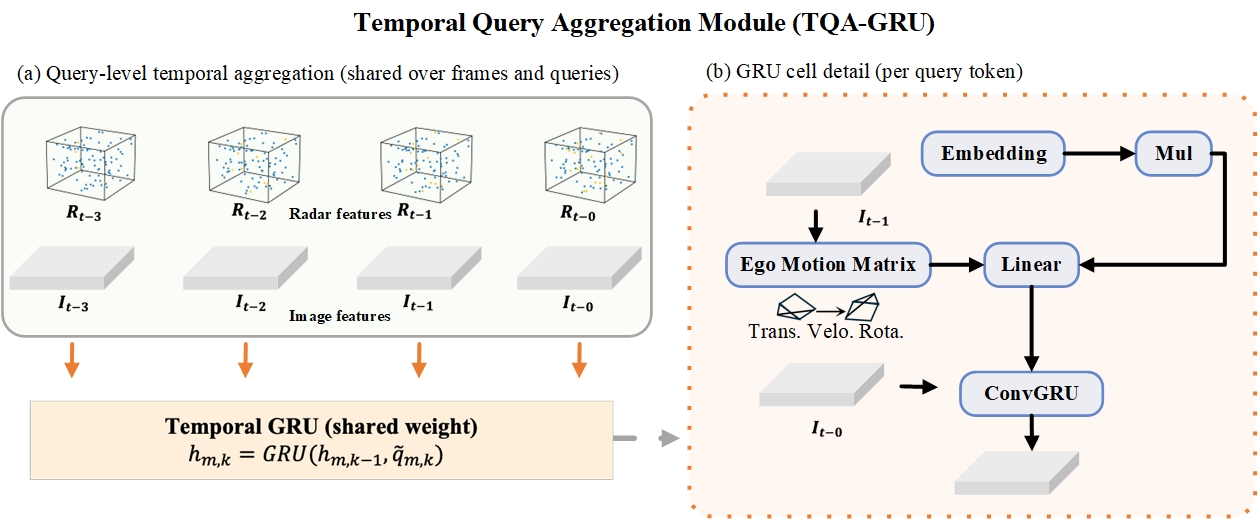}

\caption{Temporal Query Aggregation Module (TQA-GRU). (a) Query-level temporal aggregation over a window of radar and image features $\{R_{t-k}, I_{t-k}\}$, where the GRU is shared across frames and query tokens to update the hidden state $h_{m,k}$. (b) Per-query GRU cell detail: ego-motion cues (translation/velocity/rotation) are embedded and linearly projected to modulate the recurrent update (ConvGRU), enabling motion-aware temporal fusion of query representations.}

    \label{fig:TQA-GRU}
\end{figure*}
\subsection{TQA: Temporal Query Aggregation}
\label{subsec:tqa}
Water-surface observations are temporally unstable due to platform motion, wave-induced clutter, and intermittent radar returns, which can make per-frame fusion inconsistent. To improve frame-to-frame continuity, TQA-GRU aggregates fused radar--image queries over a short temporal window using a lightweight recurrent update, as shown in Fig.~\ref{fig:TQA-GRU}.
\paragraph{Temporal window and ego-motion compensation.}
We consider a temporal window of $T$ frames:
\begin{equation}
\mathcal{P}_t = \{P_{t-T+1},\ldots,P_t\},\quad
\mathcal{I}_t = \{I_{t-T+1},\ldots,I_t\}.
\end{equation}
When ego-motion estimates are available, we transform past radar frames into the current ego coordinate system:
\begin{equation}
\tilde{\mathbf{x}}_{i,\tau} = \mathbf{T}_{\tau\rightarrow t}\,\mathbf{x}_{i,\tau},\quad \tau\in[t-T+1,t],
\end{equation}
and re-encode each compensated frame so that temporal fusion is performed in a consistent reference frame.

\paragraph{Frame-wise encoding and fixed-size query representation.}
For each frame $\tau$, we compute radar and image features
\begin{equation}
\resizebox{\linewidth}{!}{$
F^{r}_{\tau}=\textsc{RadarBackbone}(\tilde{P}_{\tau}),\qquad
F^{img}_{\tau}=\textsc{ImageBackbone}(I_{\tau})
$}
\end{equation}
where the radar backbone includes the PIR Encoder and dual-stream radar modeling. 
The specific backbone instantiations (PointNet++ for radar and ResNet101 for images) are described in the experimental settings.
To obtain a fixed-size representation suitable for set prediction, we pool per-frame features into query features:
\begin{equation}
\resizebox{\linewidth}{!}{$
q^{r}_{m,\tau}=\mathrm{CrossAttn}(q_m,\ F^{r}_{\tau}),\qquad
q^{img}_{m,\tau}=\mathrm{CrossAttn}(q_m,\ F^{img}_{\tau})
$}
\end{equation}
\begin{equation}
\tilde{q}_{m,\tau}=\mathrm{MLP}_{\mathrm{fuse}}([q^{r}_{m,\tau};\,q^{img}_{m,\tau}]).
\end{equation}
This yields a temporal sequence for each query: $\{\tilde{q}_{m,t-T+1},\ldots,\tilde{q}_{m,t}\}$.

\paragraph{GRU-based temporal aggregation (TQA-GRU).}
To improve frame-to-frame continuity under platform motion and intermittent observations, we aggregate each query sequence using a lightweight GRU with time encoding $\pi(\tau)$:

\begin{equation}
h_{m,\tau}=\mathrm{GRU}\!\left(\tilde{q}_{m,\tau}+\pi(\tau),\ h_{m,\tau-1}\right),\qquad
\bar{q}_{m,t}=h_{m,t},
\end{equation}
where $h_{m,\tau}$ is the hidden state for query $m$ at time $\tau$.
The temporally aggregated query set is $\bar{Q}_t=\{\bar{q}_{m,t}\}_{m=1}^{M}$.

\paragraph{Detection head.}
Given $\bar{q}_{m,t}$, we predict target existence/class and 2D state on the water surface:
\begin{equation}
\hat{c}_m = h_{\text{cls}}(\bar{q}_{m,t}),\qquad
\hat{\mathbf{b}}_m = h_{\text{box}}(\bar{q}_{m,t}),
\end{equation}
where $\hat{\mathbf{b}}_m$ denotes the 2D water-surface state (e.g., center $(\hat{x},\hat{y})$, size $(\hat{l},\hat{w})$, heading $\hat{\theta}$).

\paragraph{Set prediction and matching.}
We adopt a set prediction formulation and match predictions to ground truth targets using bipartite matching (e.g., Hungarian algorithm) with cost

\begin{equation}
\mathcal{C} = \lambda_{\text{cls}}\mathcal{C}_{\text{cls}} + \lambda_{\text{box}}\mathcal{C}_{\text{box}},
\end{equation}
where $\mathcal{C}_{\text{cls}}$ measures classification disagreement and $\mathcal{C}_{\text{box}}$ measures box distance (e.g., $\ell_1$ on center/size plus a heading term). The matching yields positive pairs $\mathcal{M}$ between queries and targets.

\paragraph{Training objectives and temporal regularization.}
The overall loss is:
\begin{equation}
\mathcal{L}=\mathcal{L}_{\text{cls}}+\lambda_b \mathcal{L}_{\text{box}}+\lambda_t \mathcal{L}_{\text{temp}}.
\end{equation}
\emph{Classification loss} $\mathcal{L}_{\text{cls}}$ can be focal loss; \emph{box regression loss} $\mathcal{L}_{\text{box}}$ uses smooth-$\ell_1$ on the matched set $\mathcal{M}$.
To encourage stable predictions across time, we impose a temporal smoothness loss on matched targets:
\begin{equation}
\mathcal{L}_{\text{temp}}=\sum_{(m,k)\in\mathcal{M}}
\left\| \hat{\mathbf{x}}_{m,t} - \hat{\mathbf{x}}_{m,t-1} \right\|_1,\quad
\hat{\mathbf{x}}_{m,t}=(\hat{x}_{m,t},\hat{y}_{m,t}).
\end{equation}
\subsection{End-to-end forward pass.}
Putting the above components together, PhysFusion processes each frame within a temporal window using the PIR Encoder and the dual-stream radar backbone, performs radar-guided query-level fusion via RIFM, and aggregates the fused queries with TQA to obtain the temporally consistent query set $\bar{Q}_t=\{\bar{q}_{m,t}\}_{m=1}^{M}$ at time $t$. The detection head then predicts $\{(\hat{c}_m,\hat{\mathbf{b}}_m)\}_{m=1}^{M}$, and the whole framework is optimized end-to-end with set-prediction supervision. Algorithm~\ref{alg:physfusion_forward} lists the complete forward pass.
\begin{algorithm}[H]
\caption{PhysFusion Forward Pass }
\label{alg:physfusion_forward}
\begin{algorithmic}[1]
\STATE \textbf{Input:} Radar point clouds $\{P_{\tau}\}_{\tau=t-T+1}^{t}$, images $\{I_{\tau}\}_{\tau=t-T+1}^{t}$, object queries $\{q_m\}_{m=1}^{M}$, ego-motion transforms $\{\mathbf{T}_{\tau\rightarrow t}\}$ 
\STATE \textbf{Output:} Detections $\{(\hat{c}_m,\hat{\mathbf{b}}_m)\}_{m=1}^{M}$ at time $t$
\STATE Initialize query hidden states $h_{m,t-T}\gets \mathbf{0}$ for all $m=1,\dots,M$
\FOR{$\tau=t-T+1$ \textbf{to} $t$}
    \STATE \textbf{ego-motion compensation:} $\tilde{P}_{\tau}\gets \textsc{Transform}(P_{\tau},\mathbf{T}_{\tau\rightarrow t})$
    \STATE \textbf{Radar encoding (PIR + dual-stream):} $F^{r}_{\tau}\gets \textsc{RadarBackbone}(\tilde{P}_{\tau})$
    \STATE \textbf{Image encoding:} $F^{img}_{\tau}\gets \textsc{ImageBackbone}(I_{\tau})$
    \FOR{$m=1$ \textbf{to} $M$}
        \STATE \textbf{RIFM query-level fusion:}
        \STATE \hspace{0.6cm}$q^{r}_{m,\tau}\gets \textsc{CrossAttn}(q_m,\ F^{r}_{\tau})$
        \STATE \hspace{0.6cm}$q^{img}_{m,\tau}\gets \textsc{CrossAttn}(q_m,\ F^{img}_{\tau})$
        \STATE \hspace{0.6cm}$\tilde{q}_{m,\tau}\gets \textsc{MLP}_{\mathrm{fuse}}([q^{r}_{m,\tau};\,q^{img}_{m,\tau}])$
        \STATE \textbf{TQA-GRU:} $h_{m,\tau}\gets \textsc{GRU}(h_{m,\tau-1},\,\tilde{q}_{m,\tau}+\pi(\tau))$
    \ENDFOR
\ENDFOR
\FOR{$m=1$ \textbf{to} $M$}
    \STATE $\hat{c}_m\gets h_{\text{cls}}(h_{m,t})$, \quad $\hat{\mathbf{b}}_m\gets h_{\text{box}}(h_{m,t})$
\ENDFOR
\STATE \textbf{return} $\{(\hat{c}_m,\hat{\mathbf{b}}_m)\}_{m=1}^{M}$
\end{algorithmic}
\end{algorithm}

\section{Experiments}
This section presents the experimental evaluation of PhysFusion on 2D object detection tasks, along with ablation studies analyzing the impact of key components within the proposed framework.
\begin{figure}[!t]
		\centering
		\includegraphics[width=\linewidth]{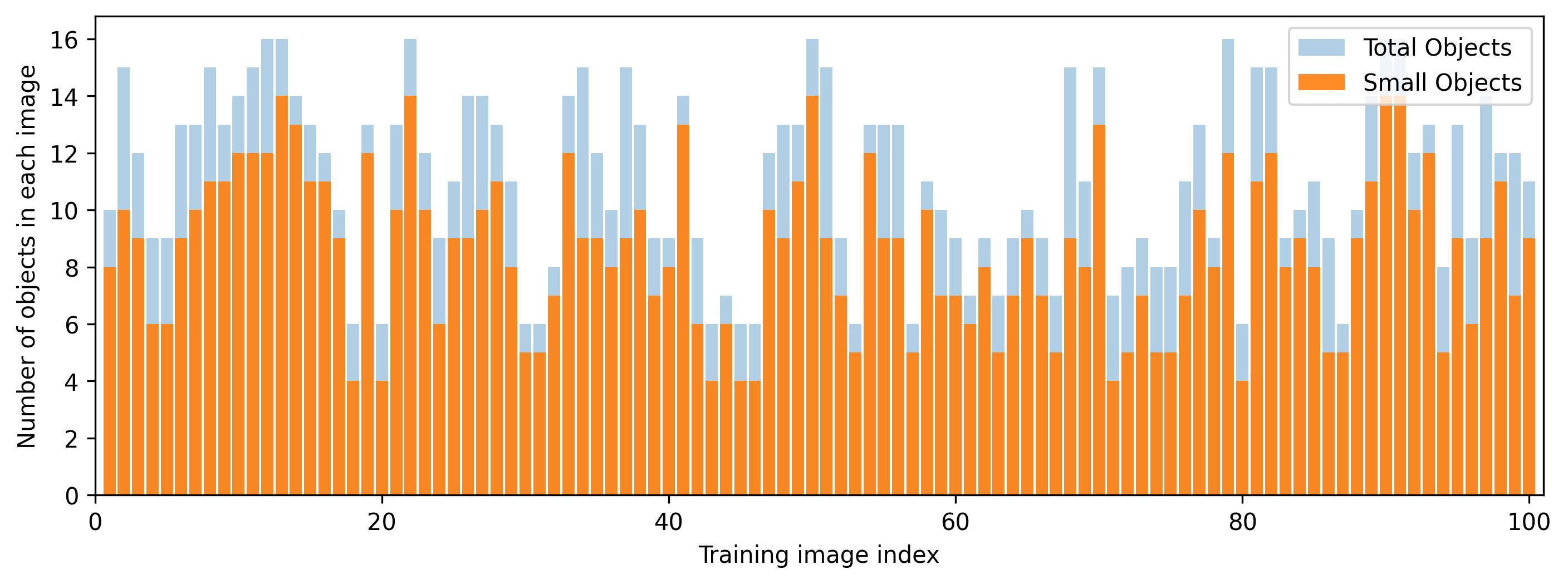}
		\caption{Per-image object statistics on the training set. 
    For each sampled image, we report the total number of labeled instances and the number of instances that satisfy a small-size criterion under the input resolution used for training.
}
		\label{fig5}
	\end{figure}
\begin{figure}[!t]
		\centering
		\includegraphics[width=\linewidth]{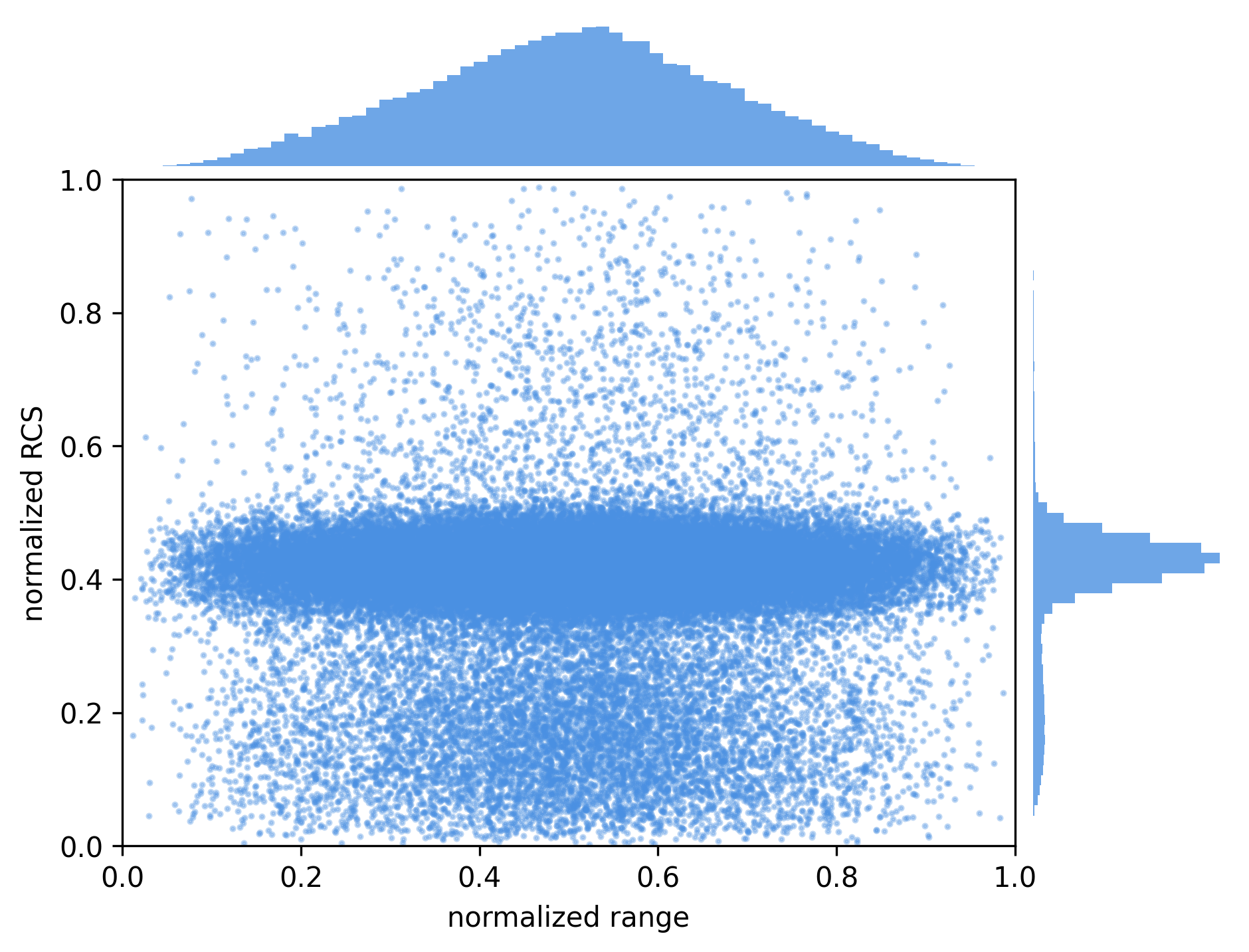}
		\caption{Statistics of radar attributes on WaterScenes. Joint distribution of normalized range and normalized RCS, showing non-uniform range coverage and a skewed long-tail tendency in reflectivity-related measurements.
}
		\label{fig6}
	\end{figure}
\subsection{Datasets and Evaluation Metrics}
\textbf{WaterScenes.} We conduct the main evaluation on WaterScenes~\cite{yao2024waterscenes}, a multi-modal benchmark for water-surface perception with synchronized RGB images and 4D millimeter-wave radar. Unless otherwise specified, we focus on 2D object detection and report mAP$_{50:95}$ and mAP$_{50}$ for accuracy, together with model complexity (Params, FLOPs) and inference speed measured on an embedded device (FPS$_\mathrm{edge}$) and a desktop GPU (FPS$_\mathrm{gpu}$). Table~\ref{tab:main_results_v2} summarizes the main results.

\textbf{FLOW.} We additionally evaluate on the FLOW dataset to examine the behavior under different sensor settings (camera-only, radar-only, and radar+camera). We report mAP$_{50}$ and mAP$_{50:95}$ in Table~\ref{tab:flow_main_results}.

\textbf{Dataset characteristics.}
Beyond reporting benchmark metrics, we summarize several dataset-level properties that are closely related to water-surface sensing. As shown in Fig.~\ref{fig5}, the number of annotated instances varies notably across images, and a non-trivial portion of instances appear at relatively small pixel scales under our training resolution. This is consistent with the prevalence of long-range observations and wide field-of-view settings during USV navigation, where appearance cues can be weak and background interference is common.

We further visualize radar attribute statistics in Fig.~\ref{fig6}. The joint distribution of normalized range and normalized RCS exhibits non-uniform range coverage and a skewed long-tail tendency in reflectivity-related measurements, suggesting that directly using raw reflectivity channels can be sensitive to scattering- and multipath-induced variations. This observation is consistent with the motivation of introducing a compact scattering prior and reliability-aware modulation in PhysFusion.

\textbf{Metrics.} Following common practice in 2D detection, mAP$_{50:95}$ averages AP over IoU thresholds from 0.50 to 0.95 with a step of 0.05, while mAP$_{50}$ reports AP at IoU=0.50. For efficiency, Params (M) and FLOPs (G) characterize model size and computation. FPS$_\mathrm{edge}$ and FPS$_\mathrm{gpu}$ are measured under a consistent runtime setting for each platform (see table footnotes).
\begin{table*}[t]
\centering
\caption{Main comparison on WaterScenes. ``C'' denotes camera, ``R'' denotes radar. $R_{1}$ and $R_{3}$ indicate using 1-frame and 3-frame radar history, respectively.}
\label{tab:main_results_v2}

\small
\setlength{\tabcolsep}{5.2pt}
\renewcommand{\arraystretch}{1.15}

\begin{threeparttable}
\begin{tabular}{l c c
                c c
                c c
                c c}
\toprule
\multirow{2}{*}{Methods} &
\multirow{2}{*}{Sensors} &
\multirow{2}{*}{Hist.\ $T$\tnote{a}} &
\multicolumn{2}{c}{Model Complexity} &
\multicolumn{2}{c}{OD (2D Detection)} &
\multicolumn{2}{c}{Speed} \\
\cmidrule(lr){4-5}\cmidrule(lr){6-7}\cmidrule(lr){8-9}
& & &
Params (M)$\downarrow$ &
FLOPs (G)$\downarrow$ &
mAP$_{50:95}$$\uparrow$ &
mAP$_{50}$$\uparrow$ &
FPS$_\mathrm{edge}$$\uparrow$\tnote{b} &
FPS$_\mathrm{gpu}$$\uparrow$\tnote{c} \\
\midrule

\multicolumn{9}{l}{\textbf{Camera-only baselines}} \\
YOLOv4-Tiny~\cite{bochkovskiy2020yolov4}             & C & 1 & 5.89 &  4.04  & 13.1 & 36.3 & \textbf{116.3}  & \textbf{342.7} \\
YOLOv7-Tiny ~\cite{wang2022yolov7}            & C & 1 & 6.03 & 33.3   & 37.3 & 65.9 &  28.9  & 120.1 \\
YOLOX-Tiny~\cite{ge2021yolox}              & C & 1 & 5.04 &  \textbf{3.79}  & 39.4 & 68.0 &  31.8  & 103.1 \\

Deformable DETR~\cite{zhu2021deformable}         & C & 1 & 39.7 & 171.2  & 56.5 & 84.0 &   10.6  &  20.3 \\
RT-DETR~\cite{Zhao_2024_CVPR}                 & C & 1 & 19.8 &  57.0  & 57.0 & 85.8 &  9.4  &  23.7 \\
YOLOv8-M~\cite{10533619}                & C & 1 & 25.7 &  78.7  &59.2 & 84.4 &  15.3  &  56.9 \\
YOLOv10-M~\cite{wang2024yolov10}               & C & 1 & 24.4 & 120.2  & 57.8 & 85.1 &  16.5  &  54.7 \\
YOLOv10-N~\cite{wang2024yolov10}              & C & 1 & 12.58 & 26.3  & 56.2 & 85.4 &  17.8  &  75.9 \\
YOLOv11-L~\cite{jocher2024ultralyticsyolo11}               & C & 1 & 23.2 &  85.9  & 57.3 & 86.1 &  23.1  &  78.5 \\
\midrule

\multicolumn{9}{l}{\textbf{Multi-modal baselines (C+R)}} \\
Achelous-MV-GDF-S0~\cite{10422325}   & C+R & 5 &  \textbf{1.6} &  -  & 51.0 & 81.1 & 15.2 & 62.8 \\
Achelous-MV-GDF-S1~\cite{10422325}   & C+R & 5 &  2.8 &  -  & 54.1& 83.5 & 15.0 & 68.4 \\
Achelous-MV-GDF-S2~\cite{10422325}   & C+R & 5 &  5.3 &  -  & 56.0& 85.5 & 15.0 & 67.0\\
YOLOv8-M + $R_1$~\cite{10533619}         & C+R & 1 & 26.7  & 84.6   & 61.5& 88.0 & 14.7 & 54.2 \\
YOLOv8-M + $R_3$~\cite{10533619}         & C+R & 3 & 27.0  & 84.8   & 62.5 & 88.8 & 14.1 & 54.2 \\
YOLOv10-M + $R_1$~\cite{wang2024yolov10}       & C+R & 1 & 24.7  & 121.9  &59.5 & 86.1 & 13.3 & 51.2 \\
YOLOv10-M + $R_3$~\cite{wang2024yolov10}       & C+R & 3 & 25.2  & 121.3  & 58.9& 87.4 & 12.7 & 51.2 \\
YOLOv11-L + $R_1$~\cite{jocher2024ultralyticsyolo11}       & C+R & 1 & 25.7  & 87.3   & 59.2 & 86.2 & 19.7 & 60.4 \\
YOLOv11-L + $R_3$~\cite{jocher2024ultralyticsyolo11}       & C+R & 3 & 25.8  & 87.7   & 60.3 & 87.7 & 18.9 & 58.3 \\
\midrule

\multicolumn{9}{l}{\textbf{Ours}} \\
PhysFusion + $R_1$      & C+R & 1 & 5.1 & 12.3   & 58.8 & 87.2 & 18.3 & 63.9 \\
PhysFusion + $R_3$      & C+R & 3 & 4.5 & 11.9   & 59.3 & 88.5 & 17.8 & 62.4 \\
PhysFusion + $R_5$      & C+R & 5 & 5.6 & 12.5   & \textbf{59.7} & \textbf{90.3} & 17.8 & 67.3 \\
\bottomrule
\end{tabular}

\begin{tablenotes}[flushleft]
\footnotesize
\item[a] $T$ indicates the radar history length used for temporal aggregation (for camera-only methods, $T=1$).
\item[b] FPS$_\mathrm{edge}$: inference speed on an embedded device (Jetson AGX Orin 32GB).
\item[c] FPS$_\mathrm{gpu}$: inference speed on a desktop GPU (NVIDIA RTX 4090).
\end{tablenotes}
\end{threeparttable}
\end{table*}
\subsection{Implementation Details}
 The overall pipeline follows Fig.~\ref{fig:main} and comprises three components: the PIR Encoder, RIFM, and TQA-GRU. At each timestamp $\tau$, the radar point cloud $P_\tau$ is first encoded by the PIR Encoder with an RCS Mapper (compact scattering prior) and a Quality Gate (point-wise reliability). The resulting point embeddings are then processed by a dual-stream radar backbone built on PointNet++~\cite{NIPS2017_d8bf84be} as the point-based local stream, together with a Transformer-based global stream equipped with SASA, producing frame-wise radar features $F^r_\tau$.

For the image branch, we employ ResNet-101~\cite{Szegedy_Ioffe_Vanhoucke_Alemi_2017} as the backbone to extract multi-scale visual features from RGB images, which are then used by RIFM for query-level radar--image fusion. The fused cross-modal queries are further aggregated over a short temporal window by TQA-GRU to obtain temporally consistent representations for set-prediction detection.

For training, we adopt the Class-Balanced Grouping and Sampling (CBGS) strategy~\cite{zhu2019class}, to alleviate class imbalance. Unless otherwise stated, all methods are trained under the same data splits and evaluation protocol. PhysFusion is trained end-to-end using AdamW~\cite{loshchilov2017decoupled} with a base learning rate of $1\times10^{-4}$, weight decay of $1\times10^{-2}$, and a batch size of 8. We use linear warmup followed by a cosine learning-rate schedule (decaying to a small minimum learning rate) for stable convergence, and enable automatic mixed-precision training with an exponential moving average (EMA) of model weights for evaluation. Data augmentations, including horizontal flipping, photometric distortion, and random scaling, are applied to RGB images. Radar points are augmented with Gaussian noise and random point dropout to improve robustness under sparse and intermittent returns.

\subsection{Main Results}
\subsubsection{Results on WaterScenes.}
Table~\ref{tab:main_results_v2} reports quantitative comparisons on WaterScenes under a unified evaluation protocol. 
We summarize the observations from three aspects: (i) accuracy--efficiency trade-off among camera-only detectors, (ii) the effect of introducing radar history in multi-modal settings, and (iii) inference behavior on edge and desktop platforms.

Camera-only methods show an accuracy--efficiency trade-off.
Lightweight detectors (e.g., YOLOv4-Tiny and YOLOX-Tiny) achieve high throughput (up to 116.3 FPS$_\mathrm{edge}$ / 342.7 FPS$_\mathrm{gpu}$ for YOLOv4-Tiny), but their detection accuracy is substantially lower (13.1 mAP$_{50:95}$ for YOLOv4-Tiny).
Transformer-based detectors provide stronger accuracy at higher computational cost and lower throughput: Deformable DETR reaches 56.5 mAP$_{50:95}$ with 171.2G FLOPs and 39.7M parameters, while RT-DETR attains the best camera-only mAP$_{50:95}$ of 57.0 with 57.0G FLOPs and 19.8M parameters, both running below 11 FPS$_\mathrm{edge}$.
YOLO-style mid/large models offer a different balance: for example, YOLOv8-M achieves 59.2 mAP$_{50:95}$ with 78.7G FLOPs, while YOLOv11-L reaches 57.3 mAP$_{50:95}$ and 86.1 mAP$_{50}$ with 23.1 FPS$_\mathrm{edge}$ and 78.5 FPS$_\mathrm{gpu}$.

\subsubsection{Multi-modal baselines: effect of radar history and fusion design.}
The multi-modal baselines (C+R) show that incorporating radar can improve detection, while the magnitude of gains depends on the fusion strategy and the history length. For example, YOLOv8-M improves from 59.2/84.4 (camera-only) to 61.5/88.0 with $R_1$ and further to 62.5/88.8 with $R_3$, at a modest reduction in embedded speed (15.3 $\rightarrow$ 14.7 $\rightarrow$ 14.1 FPS$_\mathrm{edge}$). YOLOv11-L shows a similar trend in mAP$_{50:95}$ (57.3 $\rightarrow$ 59.2 $\rightarrow$ 60.3 for camera-only, $R_1$, and $R_3$), while its mAP$_{50}$ changes more moderately (86.1 $\rightarrow$ 86.2 $\rightarrow$ 87.7). These results suggest that radar history can be beneficial, but the effectiveness is coupled with how radar observations are represented and integrated.

\begin{table}[t]
\centering
\caption{Performance comparison on the FLOW dataset under different sensor settings.}
\label{tab:flow_main_results}
\begin{tabular}{l l c c}
\hline
\textbf{Performance} & \textbf{Sensor} & \textbf{mAP$_{50}$} & \textbf{mAP$_{50:95}$} \\
\hline

\multicolumn{4}{l}{\textbf{Camera-only}}\\
YOLOv8-N~\cite{10533619}      & Camera & 75.2 & 29.9 \\
YOLOv8-S~\cite{10533619}      & Camera & 79.1 & 32.9 \\
YOLOv8-M~\cite{10533619}      & Camera & 79.5 & 33.1 \\
YOLOv8-L~\cite{10533619}      & Camera & 75.6 & 34.0 \\
YOLOv8-X~\cite{10533619}      & Camera & 79.7 & 34.3 \\
YOLOv10-L~\cite{wang2024yolov10}     & Camera & 80.7 & 44.1 \\
YOLOv11-L~\cite{jocher2024ultralyticsyolo11}     & Camera & 83.9 & 47.3 \\
RT-DETR-R18~\cite{Zhao_2024_CVPR}   & Camera & 81.7 & 45.2 \\
RT-DETR-R50~\cite{Zhao_2024_CVPR}   & Camera & 83.0 & 46.6 \\
RT-DETR-R501~\cite{Zhao_2024_CVPR}  & Camera & 84.4 & 48.2 \\
USVRT-DETR~\cite{ZHANG2025122926}    & Camera & 85.6 & 49.6 \\
\hline

\multicolumn{4}{l}{\textbf{Radar-only}}\\
YOLOv8-N~\cite{10533619}      & Radar  & 72.5 & 39.4 \\
YOLOv8-S~\cite{10533619}      & Radar  & 69.1 & 38.8 \\
YOLOv8-M~\cite{10533619}      & Radar  & 74.4 & 42.1 \\
YOLOv8-L~\cite{10533619}      & Radar  & 71.0 & 39.3 \\
YOLOv8-X~\cite{10533619}      & Radar  & 70.9 & 39.7 \\
YOLOv10-L~\cite{wang2024yolov10}     & Radar  & 70.6 & 39.5 \\
YOLOv11-L~\cite{jocher2024ultralyticsyolo11}     & Radar  & 73.5 & 41.2 \\
\hline

\multicolumn{4}{l}{\textbf{Radar + Camera}}\\
RCFNet~\cite{10446880}        & Radar + Camera & 93.2 & 44.7 \\
\hline
\textbf{PhysFusion (ours)} & \textbf{Radar + Camera} & \textbf{94.8} & \textbf{46.2} \\
\hline
\end{tabular}
\end{table}
\begin{figure}[!t]
		\centering
		\includegraphics[width=\linewidth]{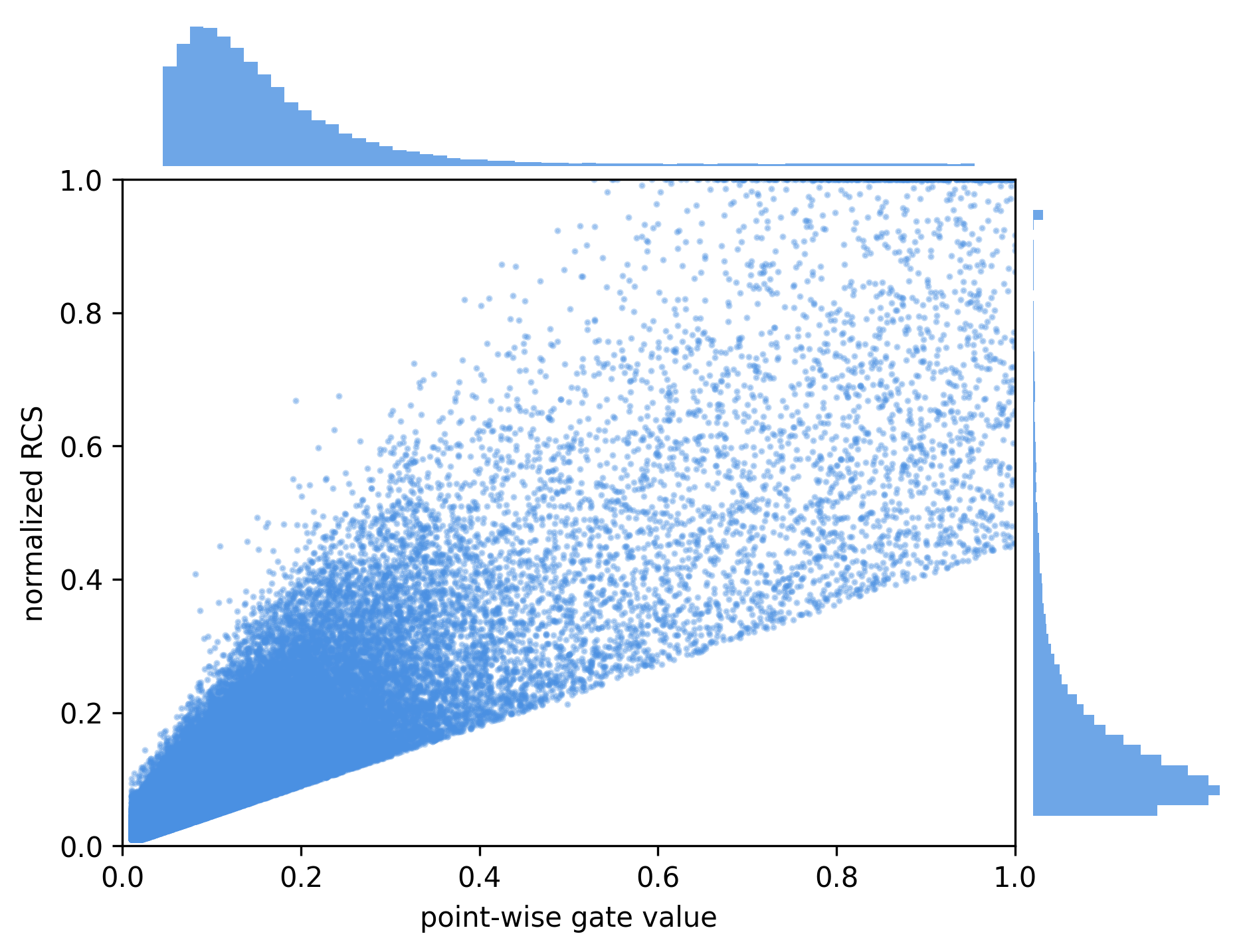}
		\caption{Per-image object statistics on the training set. 
    For each sampled image, we report the total number of labeled instances and the number of instances that satisfy a small-size criterion under the input resolution used for training.
}
		\label{fig:gate_rcs}
	\end{figure}
\subsubsection{PhysFusion.} PhysFusion achieves 58.8/87.2 (mAP$_{50:95}$/mAP$_{50}$) with $R_1$, 59.3/88.5 with $R_3$, and 59.7/90.3 with $R_5$, while keeping a compact computation profile (11.9--12.5G FLOPs and 4.5--5.6M parameters) and practical speed (17.8--18.3 FPS$_\mathrm{edge}$ and 62.4--67.3 FPS$_\mathrm{gpu}$). Compared with the strongest multi-modal baseline in terms of mAP$_{50:95}$ (YOLOv8-M+$R_3$: 62.5), PhysFusion provides a higher mAP$_{50}$ (90.3 vs.\ 88.8) with substantially lower FLOPs (12.5G vs.\ 84.8G) and fewer parameters (5.6M vs.\ 27.0M), indicating a favorable accuracy--efficiency balance under the radar+camera setting.

\subsubsection{Impact of radar history length ($R_1$/$R_3$/$R_5$).}
Increasing the radar history length consistently improves PhysFusion performance in Table~\ref{tab:main_results_v2}: mAP$_{50:95}$ increases from 58.8 ($R_1$) to 59.3 ($R_3$) and 59.7 ($R_5$), while mAP$_{50}$ increases from 87.2 to 88.5 and 90.3. The corresponding runtime change is small on the embedded device (18.3 $\rightarrow$ 17.8 FPS$_\mathrm{edge}$), and the desktop speed remains within a similar range (63.9--67.3 FPS$_\mathrm{gpu}$). This trend is consistent with the motivation of leveraging short-term temporal context to mitigate intermittent radar returns.
\begin{table*}[t]
\centering
\caption{Ablation study of PhysFusion components on WaterScenes. ``$\checkmark$'' indicates that the corresponding module is enabled. Here the RCS prior module includes both the scattering prior $s$ and the point-wise quality gate $g$. The reported values follow the same protocol as Table~\ref{tab:main_results_v2}.}
\label{tab:ablation_physfusion}
\small
\setlength{\tabcolsep}{5.2pt}
\renewcommand{\arraystretch}{1.12}
\begin{threeparttable}
\begin{tabular}{c c c c c c c c}
\toprule
Exp. &
 PIR Encode ($s{+}g$) &
RIFM &
TQA-GRU &
mAP$_{50}$ $\uparrow$ &
mAP$_{50:95}$ $\uparrow$ &
Params (M) $\downarrow$ &
FLOPs (G) $\downarrow$ \\

\midrule
1 & - & - & - & 85.1 & 56.2 & 5.4 & 11.7 \\
2 & \checkmark & - & - & 85.3 & 57.6 & 5.5 & 10.2 \\
3 & - & \checkmark & - & 85.9 & 58.2 & 6.8 & 12.9 \\
4 & - & - & \checkmark & 86.5 & 58.6 & 6.1 & 14.3 \\
5 & \checkmark & \checkmark & - & 87.9 & 59.1 & 6.1 & 14.8 \\
6 & \checkmark & - & \checkmark & 88.7 & 58.5 & 5.8 & 13.0 \\
7 & - & \checkmark & \checkmark & 88.2 & 58.0 & 5.9 & 12.7 \\
8 & \checkmark & \checkmark & \checkmark & 90.3 & 59.7 & 5.6 & 12.5 \\
\bottomrule
\end{tabular}
\begin{tablenotes}[flushleft]
\footnotesize
\item \textbf{Note:} The ``RCS Prior'' module includes both $s$ and the quality gate $g$. If you would like to keep the original ablation numbers exactly unchanged, simply replace the placeholder rows (Exp.\#5--\#7) with your measured values while keeping the same module toggles.
\end{tablenotes}
\end{threeparttable}
\end{table*}

     \begin{figure*}[ht]

        \includegraphics[width=\linewidth]{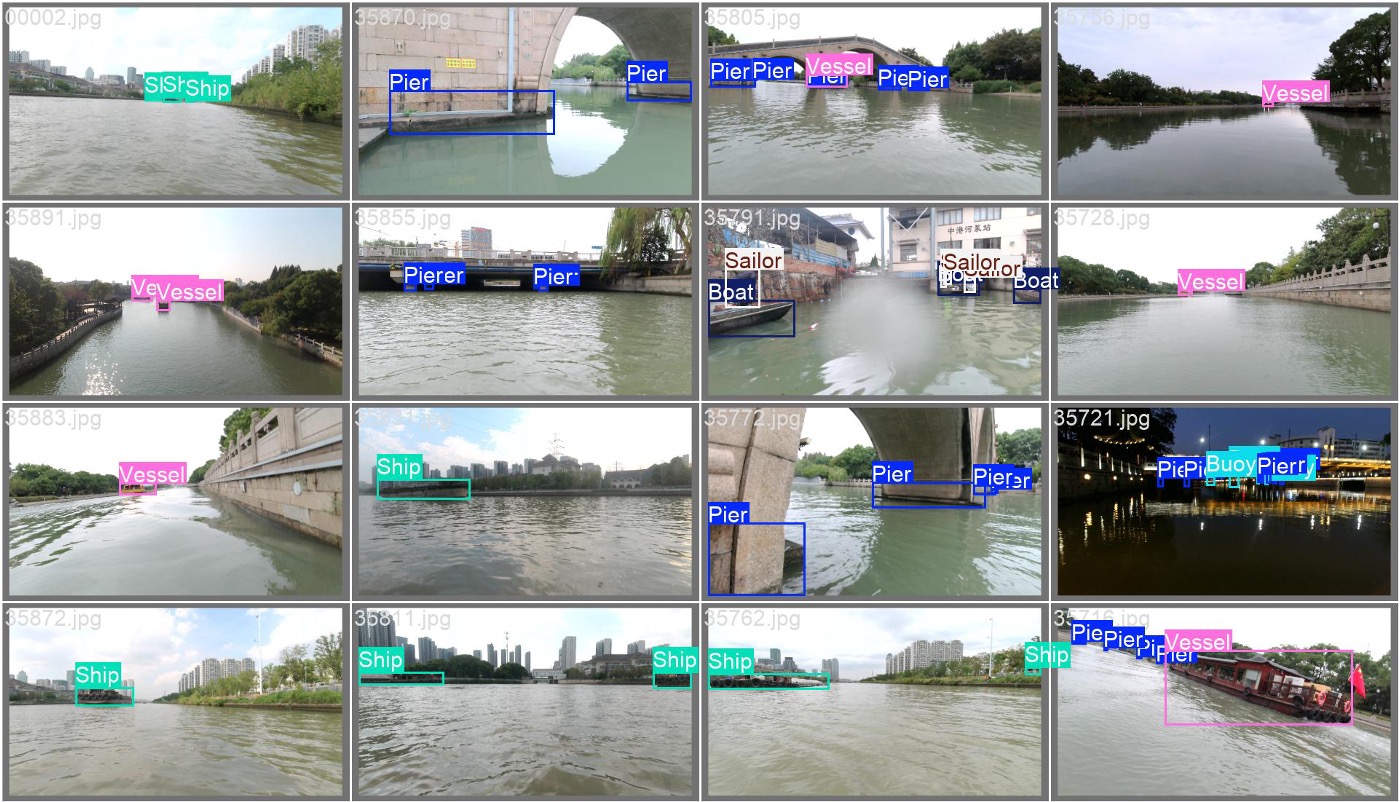}
    \caption{Qualitative detection results of the proposed PhysFusion framework on the WaterScenes dataset. The model successfully identifies water-surface objects under diverse conditions, including partial occlusion, background clutter, and varying illumination levels.}
    \label{fig:visual1}
\end{figure*}

\subsubsection{Results on FLOW.}
Table~\ref{tab:flow_main_results} evaluates camera-only, radar-only, and radar+camera settings on FLOW.
Camera-only results show increasing accuracy with stronger backbones (e.g., 47.3 mAP$_{50:95}$ for YOLOv11-L and 49.6 for USVRT-DETR), establishing the upper range achievable with vision cues alone on this dataset.
Radar-only baselines exhibit different behavior: their mAP$_{50:95}$ values are competitive for certain model sizes (e.g., YOLOv8-M Radar: 42.1), indicating that radar provides complementary detection cues despite sparsity.
Under radar+camera fusion, PhysFusion achieves 94.8 mAP$_{50}$ and 46.2 mAP$_{50:95}$, outperforming the radar+camera baseline RCFNet (93.2 mAP$_{50}$ and 44.7 mAP$_{50:95}$).
The improvement is more pronounced on mAP$_{50:95}$ (+1.5), suggesting that the proposed fusion and temporal query aggregation contribute to more consistent localization across IoU thresholds rather than only boosting loose-match detection.

\subsection{Ablation Study}
\label{sec:ablation}

\subsubsection{Quality Gate behavior analysis (PIR Encoder).}
To better interpret what the point-wise Quality Gate learns, we visualize the relationship between the gate output and radar reflectivity.
Fig.~\ref{fig:gate_rcs} plots the predicted reliability score $g\in[0,1]$ against normalized RCS with marginal histograms.
Two patterns are apparent: (i) points with similar RCS values can receive substantially different $g$ scores, suggesting that the gate is not a simple monotonic rescaling or thresholding of reflectivity; and (ii) a subset of points with extreme RCS values tends to be assigned lower $g$, which is consistent with the presence of heavy-tailed reflectivity outliers in maritime radar returns (e.g., clutter, multipath, and water-surface scattering).
Overall, the visualization supports the role of the Quality Gate as a reliability-aware filtering signal within the PIR Encoder, complementing the scattering prior produced by the RCS Mapper before subsequent feature aggregation.

\subsubsection{Component ablations on WaterScenes (PIR Encoder and global radar modeling).}
We evaluate the contributions of key components under the same data split and evaluation protocol as the main experiments.
Table~\ref{tab:ablation_physfusion} reports a progressive ablation where Exp.~1 serves as the baseline, and Exp.~2--8 enable: (i) the PIR Encoder via the RCS prior ($s{+}g$), (ii) SASA as the Scattering-Aware Self-Attention strategy for global radar modeling, and (iii) their combinations. Unless otherwise stated, all results are taken directly from Table~\ref{tab:ablation_physfusion}.

\subsubsection{Baseline (Exp.~1).}
Exp.~1 disables the PIR Encoder (i.e., no RCS prior) and SASA, yielding 56.2 mAP$_{50:95}$ and 85.1 mAP$_{50}$ with 5.4M parameters and 11.7G FLOPs.

\subsubsection{Effect of the PIR Encoder (RCS prior $s{+}g$).}
Enabling the PIR Encoder (Exp.~2) improves mAP$_{50:95}$ from 56.2 to 57.6 and mAP$_{50}$ from 85.1 to 85.3.
This change also reduces FLOPs from 11.7G to 10.2G (Params: 5.4M $\rightarrow$ 5.5M), indicating that the scattering prior and point-wise reliability modulation can provide a more effective radar representation under the same evaluation protocol.

\subsubsection{Effect of SASA for global radar modeling.}
When SASA is enabled alone (Exp.~3), performance increases to 58.2 mAP$_{50:95}$ and 85.9 mAP$_{50}$, with increased computation (Params: 5.4M $\rightarrow$ 6.8M; FLOPs: 11.7G $\rightarrow$ 12.9G). This is consistent with the role of SASA in introducing a scattering-aware distance-decayed bias for global reasoning over sparse radar points.

\subsubsection{Module combinations and full configuration.}
Combining the PIR Encoder with SASA (Exp.~5) further improves accuracy to 59.1 mAP$_{50:95}$ and 87.9 mAP$_{50}$ (6.1M Params, 14.8G FLOPs), suggesting that a scattering prior with reliability modulation and scattering-aware global attention can be complementary. Adding both components together with the remaining design choices in the full configuration (Exp.~8) yields the strongest performance in Table~\ref{tab:ablation_physfusion}, achieving 59.7 mAP$_{50:95}$ and 90.3 mAP$_{50}$ with 5.6M parameters and 12.5G FLOPs.

\subsubsection{Ablation on the Quality Gate $g$ and temporal query aggregation (TQA-GRU) on FLOW.}
In addition to the radar-encoder ablations above, Table~\ref{tab:ablation_gate_tqa} isolates the effects of the point-wise gate $g$ and the Temporal Query Aggregation module (TQA-GRU) on FLOW. Starting from the single-frame setting without $g$ and TQA-GRU (ID-(1)), enabling only $g$ (ID-(2)) improves mAP$_{50}$ from 89.5 to 90.8 and mAP$_{50:95}$ from 44.1 to 44.2. Enabling only TQA-GRU under $T{=}1$ (ID-(3)) yields 91.3 mAP$_{50}$ and 44.6 mAP$_{50:95}$, and increasing the temporal window to $T{=}3$ (ID-(4)) further improves to 92.1/45.0. Finally, combining $g$ and TQA-GRU achieves the highest performance: 93.4/45.7 for $T{=}1$ (ID-(5)) and 94.8/46.2 for $T{=}3$ (ID-(6)). These results suggest that point-wise reliability filtering and query-level temporal aggregation provide complementary gains under intermittent radar observations.

\begin{table}[t]
\centering
\caption{Ablation on the Quality Gate $g$ and the Temporal Query Aggregation (TQA-GRU) on FLOW. $R_1$/$R_3$ indicate using 1/3-frame radar history for temporal aggregation.}
\label{tab:ablation_gate_tqa}

\small
\setlength{\tabcolsep}{4.8pt}
\renewcommand{\arraystretch}{1.15}

\begin{threeparttable}
\begin{tabular}{l c c c c c}
\toprule
ID & Gate $g$ & TQA-GRU & Hist.\ $T$ & mAP$_{50:95}$ $\uparrow$ & mAP$_{50}$ $\uparrow$ \\
\midrule
(1) &  &  & 1 & 44.1 & 89.5 \\
(2) & \cmark &  & 1 & 44.2 & 90.8 \\
(3) &  & \cmark & 1 & 44.6 & 91.3 \\
(4) &  & \cmark & 3 & 45.0 & 92.1 \\
(5) & \cmark & \cmark & 1 & 45.7 & 93.4 \\
(6) & \cmark & \cmark & 3 & 46.2 & 94.8 \\
\bottomrule
\end{tabular}

\begin{tablenotes}[flushleft]
\footnotesize
\item \textbf{ID-(1)} disables both the point-wise gate and temporal aggregation (single-frame setting).
\item \textbf{Gate $g$}: point-wise reliability filtering is applied only to the local point-based stream.
\item \textbf{w/o TQA-GRU}: no temporal aggregation; inference is performed per-frame (unless otherwise specified).
\end{tablenotes}
\end{threeparttable}
\end{table}
\begin{figure*}[h]
		\centering\
  \includegraphics[width=1.0\textwidth]{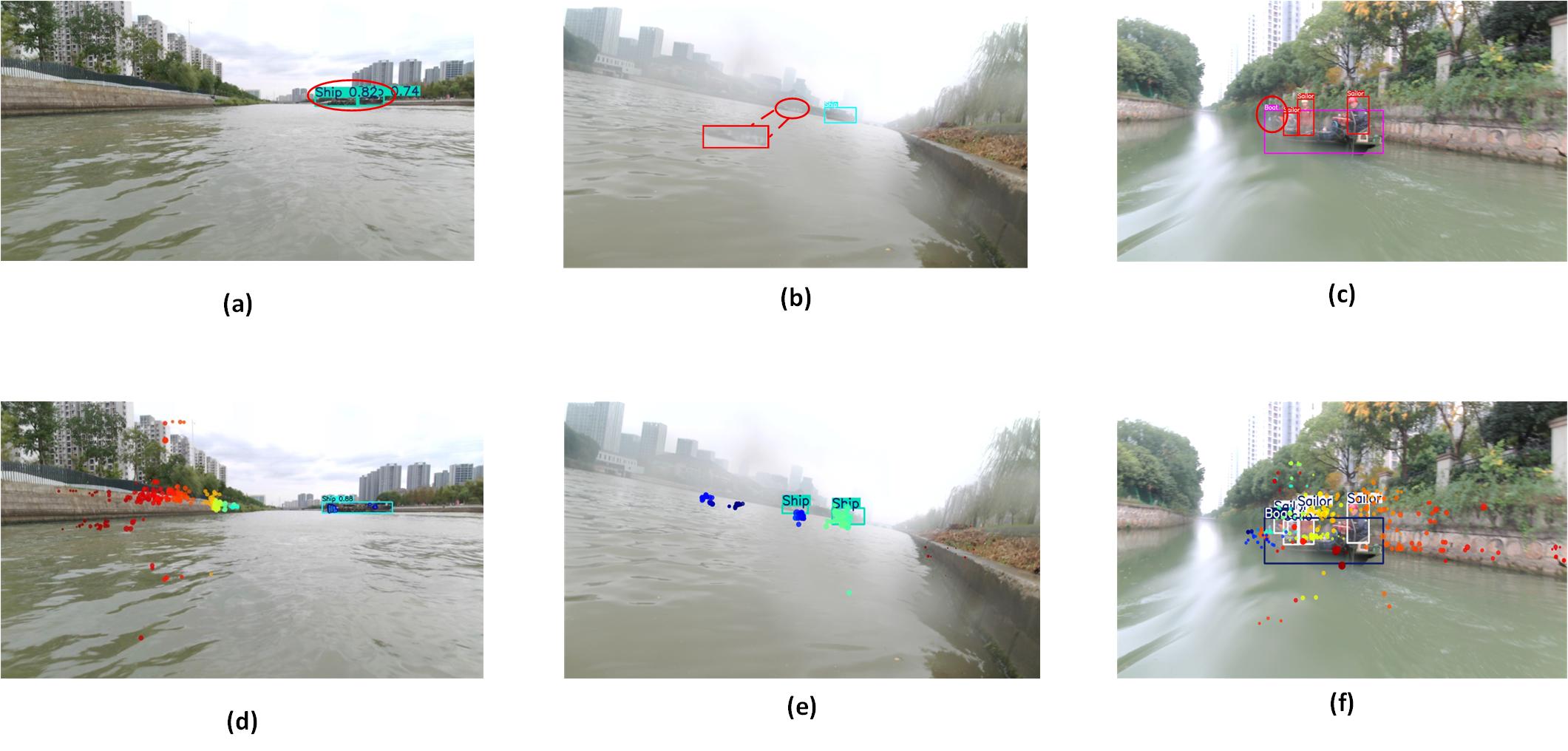}
\caption{Comparison between vision-only and radar-image fusion detection results under challenging water-surface conditions. 
Subfigures (a)--(c) show detection results from the vision-only variant PhysFusion-C, while (d)--(f) are results from the full multi-modal PhysFusion. Red ellipses in (a)--(c) indicate failure cases such as missed or false detections. }
		\label{fig:visual2}
	\end{figure*}
\subsection{Visualization Results}
\label{sec:vis}
We provide qualitative results to complement the quantitative comparisons and to illustrate how the proposed PysFusion behave under challenging water-surface conditions.

\subsubsection{Overall qualitative results.} Fig.~\ref{fig:visual1} presents representative detections produced by PhysFusion across diverse scenes, including open water, near-shore channels, bridge/riverbank structures, and night-time illumination. The examples cover multiple categories (e.g., \emph{Ship}, \emph{Boat}, \emph{Vessel}, \emph{Pier}, \emph{Buoy}, and \emph{Sailor}) and include cases with background reflections, shoreline clutter, and partial occlusion. These results indicate that fused radar--image queries can produce stable predictions across varying illumination conditions while maintaining instance separation in cluttered regions.
	\begin{figure*}[h]
		\centering\
  \includegraphics[width=1.0\textwidth]{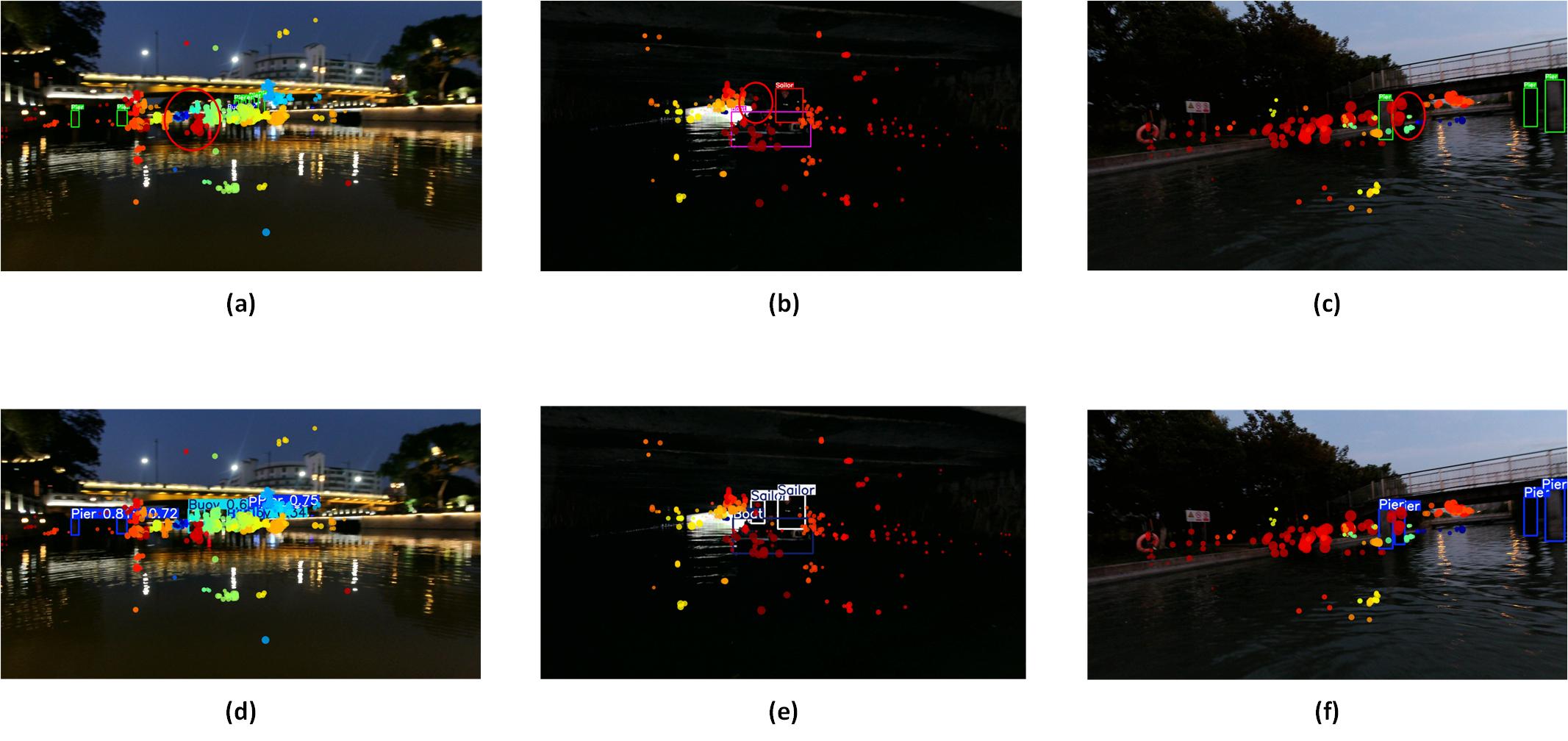}
\caption{Qualitative comparison between multi-modal YOLOv10-L (top row, a–c) and the proposed PhysFusion (bottom row, d–f) under low-light and low-visibility conditions. 
Subfigures (a)–(c) show detection results from the YOLOX-L-based multi-modal detector, while (d)–(f) present results from the full PhysFusion model. 
Red ellipses in (a)–(c) indicate missed or false detections. }
		\label{fig:visual3}
	\end{figure*}   
\subsubsection{Vision-only vs.\ radar--image PhysFusion.}
Fig.~\ref{fig:visual2} compares the vision-only variant (PhysFusion-C, (a)--(c)) with the full radar--image PhysFusion ((d)--(f)).
The red ellipses in (a)--(c) illustrate typical failure cases under vision-only inference, including missed detections and false positives caused by water-surface reflections or background structures.
With the radar branch enabled, PhysFusion leverages radar evidence and the reliability gate to reduce the influence of unstable returns before fusion, recovering detections that are missed in the vision-only setting and yielding more consistent localization in cluttered or low-contrast regions.

\subsubsection{Comparison to a multi-modal baseline under low visibility.}
Fig.~\ref{fig:visual3} further compares PhysFusion with a multi-modal YOLOv10-L baseline under low-light and low-visibility conditions. The baseline results (a)--(c) show missed targets and less precise localization when appearance cues are weak and the background contains strong reflections and sensor noise. In contrast, PhysFusion (d)--(f) leverages: (i) RIFM, where semantically enriched radar features guide query-level fusion by attending to multi-scale image features, and (ii) TQA-GRU, which aggregates fused queries over a short temporal window to reduce frame-to-frame fluctuations under intermittent radar returns. In these examples, this combination helps retain detections for small or distant targets and improves localization consistency under degraded visibility.

\section{Conclusions}
In this paper, we presented PhysFusion, a physics-informed radar--image detection framework tailored for water-surface perception. The proposed approach targets practical challenges in USV scenarios, including non-stationary wave clutter, specular reflections, long-range observations with weak visual cues, and sparse/intermittent radar returns.

PhysFusion is built on three key components. First, we introduce a Physics-Informed Radar Encoder (PIR Encoder) that combines an RCS Mapper with a Quality Gate to construct a compact scattering prior and to estimate point-wise reliability, improving radar feature robustness under heavy-tailed reflectivity variations and unstable returns. Second, we propose a Radar-guided Interactive Fusion Module (RIFM) that performs query-level radar--image fusion, enabling adaptive cross-modal interaction between semantically enriched radar queries and multi-scale visual features; the radar branch further adopts SASA for global modeling of sparse radar points. Third, we employ Temporal Query Aggregation (TQA), implemented with a lightweight GRU (TQA-GRU), to aggregate frame-wise fused queries over a short temporal window and improve frame-to-frame consistency under intermittent observations.

Experimental results on WaterScenes and FLOW show that PhysFusion attains favorable detection performance with moderate model complexity and practical inference speed. These results validate the effectiveness of PhysFusion for practical deployment in USV applications such as autonomous navigation, surface monitoring, and obstacle avoidance.

Several directions remain for future work. First, more explicit motion modeling (e.g., tracking-aware or state-space temporal aggregation) may further improve robustness under rapid platform dynamics and wave-induced occlusions. Second, extending the framework to incorporate additional sensing modalities (e.g., thermal imaging or sonar) could improve performance under extreme visibility degradation. Third, adapting PhysFusion to broader tasks such as multi-object tracking or instance segmentation may further expand its applicability for maritime autonomy.

While PhysFusion shows promising results, several directions remain open for further exploration: (1) Incorporating temporal cues from radar sequences could further enhance detection consistency in dynamic or wave-disturbed environments; (2) Integrating additional sensing modalities, such as sonar or thermal imaging, may improve robustness under extreme weather or lighting conditions; (3) Extending PhysFusion to support broader perception tasks (e.g., semantic segmentation, tracking, or panoptic perception) may further strengthen its applicability in complex maritime environments.

\section*{Acknowledgments}
This work was supported by the National Science and Technology Major Project under grant number 2022ZD0116409.


\vfill
\bibliographystyle{IEEEtran}
\bibliography{IEEEabrv,ref}
\end{document}